\documentclass[letterpaper, 10 pt, conference]{ieeeconf}  %

\IEEEoverridecommandlockouts                              %

\usepackage{adjustbox}
\usepackage[ruled, vlined]{algorithm2e}
\usepackage{amsmath}
\usepackage{amssymb}
\usepackage{array}  %
\usepackage{authblk}
\usepackage[accsupp]{axessibility}  %
\usepackage{booktabs}
\usepackage{capt-of}
\makeatletter
\let\NAT@parse\undefined
\makeatother
\usepackage{cite}
\usepackage[font={small}]{caption}
\usepackage{colortbl}
\usepackage{epsfig}
\usepackage{inconsolata}  %
\usepackage{graphicx}
\usepackage{listings}
\usepackage{multicol}
\usepackage{multirow}
\usepackage{paralist}
\usepackage{pifont}
\usepackage{ragged2e} %
\usepackage{times}
\usepackage{tabularx}
\usepackage{xcolor}
\usepackage{svg}

\definecolor{flodarkpurple}{rgb}{0.288,0.1196,0.7}

\definecolor{amber}{rgb}{1.0, 0.75, 0.0}

\usepackage[backref=page,breaklinks,colorlinks,bookmarks,citecolor=flodarkpurple]{hyperref}

\newcommand{\coolname}{\textit{ConceptGraphs}}
\newcommand{\coolnamedetect}{\coolname-\textit{Detector}}

\newcommand{\coolabbrdetect}{\textit{CG-D}}
\newcommand{\webpage}{https://concept-graphs.github.io/}

\newcommand{\authorhref}[3][flodarkpurple]{\href{#2}{\color{#1}{#3}}}

\title{\Large \bf
\coolname: Open-Vocabulary 3D Scene Graphs for Perception and Planning \\
\vspace{0.30em}
\large{\href{\webpage}{\color{violet}{\texttt{\webpage}}}}
}

\author{
\authorhref{https://georgegu1997.github.io/}{Qiao Gu}$^{\dagger*1}$, 
\authorhref{https://www.alihkw.com/}{Ali Kuwajerwala}$^{\dagger*2}$, 
\authorhref{https://sachamorin.github.io/}{Sacha Morin}$^{*2}$, 
\authorhref{https://krrish94.github.io/}{Krishna Murthy Jatavallabhula}$^{*3}$, \\
\authorhref{https://bipashasen.github.io/}{Bipasha Sen}$^{2}$,
\authorhref{https://skymanaditya1.github.io/}{Aditya Agarwal}$^{2}$, 
\authorhref{https://www.jhuapl.edu/work/our-organization/research-and-exploratory-development/red-staff-directory/corban-rivera}{Corban Rivera}$^5$,
\authorhref{https://scholar.google.com/citations?user=92bmh84AAAAJ}{William Paul}$^5$,
\authorhref{https://mila.quebec/en/person/kirsty-ellis/}{Kirsty Ellis}$^{2}$, \\
\authorhref{https://engineering.jhu.edu/faculty/rama-chellappa/}{Rama Chellappa}$^6$,
\authorhref{https://people.csail.mit.edu/ganchuang/}{Chuang Gan}$^7$,
\authorhref{https://celsodemelo.net/}{Celso Miguel de Melo}$^{4}$, \\
\authorhref{http://web.mit.edu/cocosci/josh.html}{Joshua B. Tenenbaum}$^{3}$,
\authorhref{https://groups.csail.mit.edu/vision/torralbalab/}{Antonio Torralba}$^{3}$,
\authorhref{http://www.cs.toronto.edu//~florian/}{Florian Shkurti}$^{1}$
\authorhref{http://liampaull.ca/}{Liam Paull}$^{2}$,
\\
$^{1}$\href{https://robotics.utoronto.ca/}{University of Toronto}, 
$^{2}$\href{https://montrealrobotics.ca/}{Université de Montréal}, 
$^{3}$\href{https://www.mit.edu/}{MIT},
$^{4}$\href{https://www.arl.army.mil/}{DEVCOM ARL},
$^{5}$\href{https://www.jhuapl.edu/}{JHU APL},
$^{6}$\href{https://www.jhu.edu/}{JHU},
$^{7}$\href{https://umass.edu/}{UMass}
\thanks{$\dagger$Project Lead  *Equal Contribution}
}

\begin{document}

\makeatletter
\let\@oldmaketitle\@maketitle
\renewcommand{\@maketitle}{\@oldmaketitle
\centering
\vspace{-0.5em}
\includegraphics[width=\linewidth, trim={1.1cm 0 0.8cm 0}]{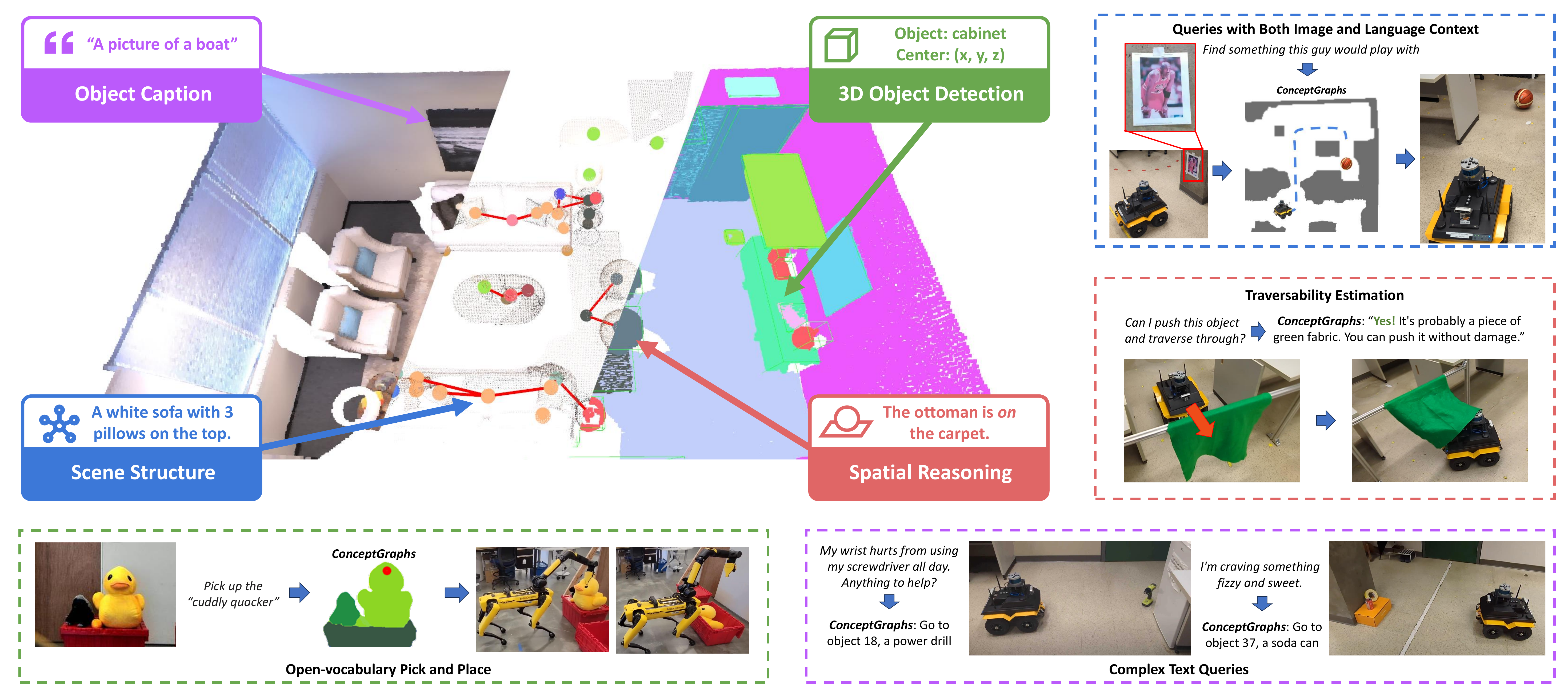}
\vspace{-1.5em}
\captionof{figure}{
\textbf{\coolname} builds open-vocabulary 3D scene graphs. We (a) design an object-based mapping system that only assumes class-agnostic instance masks and fuses them to 3D, (b) interprets and extracts language tags for each mapped instance leveraging large vision-language models, and (c) builds a graph of object spatial relationships by leveraging priors encoded in large language models. 
The object-centric nature of \coolname{} allows easy map maintenance and promotes scalability, and the graph structure provides relational information within the scene.
Furthermore, our scene graph representations are easily mapped to natural language formats to interface with LLMs, enabling them 
to answer complex scene queries and granting robots access to useful facts about surrounding objects, such as traversability and utility.
We implement and demonstrate \coolname{} on a number of real-world robotics tasks across wheeled and legged mobile robot platforms. (\href{https://concept-graphs.github.io/}{Webpage}) (\href{https://youtu.be/mRhNkQwRYnc}{Explainer Video})
}
\label{fig:splash}
\vspace{-0.98em}
}
\makeatother

\maketitle
\thispagestyle{empty}
\pagestyle{empty}

\begin{abstract}

For robots to perform a wide variety of tasks, they require a 3D representation of the world that is semantically rich, yet compact and efficient for task-driven perception and planning.
Recent approaches have attempted to leverage features from large vision-language models to encode semantics in 3D representations. 
However, these approaches tend to produce maps with per-point feature vectors, which do not scale well in larger environments, nor do they contain semantic spatial relationships between entities in the environment, which are useful for downstream planning. 
In this work, we propose \coolname{}, an open-vocabulary graph-structured representation for 3D scenes. 
\coolname~is built by leveraging 2D foundation models and fusing their output to 3D by multi-view association. 
The resulting representations generalize to novel semantic classes, without the need to collect large 3D datasets or finetune models. 
We demonstrate the utility of this representation through a number of downstream planning tasks that are specified through abstract (language) prompts and require complex reasoning over spatial and semantic concepts. To explore the full scope of our experiments and results, we encourage readers to visit our \href{https://concept-graphs.github.io/}{project webpage}.

\end{abstract}

\setcounter{figure}{1} %

\section{Introduction}
\label{sec:intro}

Scene representation is one of the key design choices that can facilitate downstream planning for a variety of tasks, including mobility and manipulation. 
Robots need to build these representations online from onboard sensors as they navigate through an environment. 
For efficient execution of complex tasks
such representations should be: \underline{\emph{scalable and efficient to maintain}}, as the volume of the scene and the duration of the robot's operation increases;  \underline{\emph{open-vocabulary}}, not limited to making inferences about a set of concepts that is predefined at training time, but capable of handling new objects and concepts at inference time; and \underline{\emph{with a flexible level of detail}} to enable planning over a range of tasks, from ones that require dense geometric information for mobility and manipulation, to ones that need abstract semantic information and object-level affordance information for task planning. We propose \textbf{\coolname{}}, a 3D scene representation method for robot perception and planning that satisfies all the above requirements.

\subsection{Related Work}

\textbf{Closed-vocabulary semantic mapping in 3D.}
Early works reconstruct the 3D map through online algorithms like simultaneous localization and mapping (SLAM)~\cite{newcombe2011kinectfusion, whelan2015elasticfusion, wen2023bundlesdf, sucar2021imap, zhu2022niceslam} or offline methods like structure-from-motion (SfM)~\cite{schonberger2016sfm, schonberger2016pixelwise}. 
Aside from reconstructing 3D geometry, recent works also use deep learning-based object detection and segmentation models to reconstruct the 3D scene representations with dense semantic mapping~\cite{mccormac2017semanticfusion, runz2018maskfusion, mccormac2018fusion++, narita2019panopticfusion} or object-level decomposition~\cite{Qian2022pocd, qian2023povslam, li2021odam, zins2022oaslam}. 
While these methods achieve impressive results in mapping semantic information to 3D, they are closed-vocabulary and their applicability is limited to object categories annotated in their training datasets.

\textbf{3D scene representations using foundation models.} 
There have been significant recent efforts~\cite{liu20233dovs, conceptfusion, peng2023openscene, ding2023lowis3d, ding2023pla, zhang2023clipfo3d, tschernezki2022n3f, kobayashi2022ffd, clipfields, tsagkas2023vlfields, lerf2023, huang23avlmaps, shen2023distilled, Engelmann2023openreno, mazur2023feature} focused on building 3D representations by leveraging \textit{foundation models} - large, powerful models that capture a diverse set of concepts and accomplish a wide range of tasks~\cite{clip, gpt4, kirillov2023sam, liu2023grounding, stablediffusion}. 
Such models have excelled in tackling open-vocabulary challenges in 2D vision.
However, they require an ``internet-scale" of training data, and no 3D datasets exist yet of a comparable size. 
Recent works have therefore attempted to \textit{ground} the 2D representations produced by image and language foundation models to the 3D world and show impressive results on open-vocabulary tasks, including language-guided object grounding~\cite{conceptfusion, clipfields, peng2023openscene, lerf2023,hong20223d}, 3D reasoning~\cite{hong20233d,hong20233d-llm}, robot manipulation~\cite{shridar2021cliport, sharma2023llerftogo} and navigation~\cite{gadre2022cow, shah2022lmnav}. 
These approaches project dense per-pixel features from images to 3D to build explicit representations such as pointclouds~\cite{conceptfusion, peng2023openscene, ding2023lowis3d, ding2023pla, zhang2023clipfo3d} or implicit neural representations~\cite{liu20233dovs, tschernezki2022n3f, kobayashi2022ffd, clipfields, tsagkas2023vlfields, lerf2023, huang23avlmaps, shen2023distilled, Engelmann2023openreno, mazur2023feature}.

However, such methods have two key limitations. 
First, assigning every point a semantic feature vector is highly redundant and consumes more memory than necessary, greatly limiting scalability to large scenes.
Second, these dense representations do not admit an easy decomposition -- this lack of structure makes them less amenable to dynamic updates to the map (crucial for robotics).

\textbf{3D scene graphs.}
3D scene graphs (3DSGs) address the second limitation by compactly and efficiently describing scenes with graph structures, with nodes representing objects and edges encoding inter-object relationships~\cite{Fisher2011scenegraph, gay2019visual, Armeni20193DSG, kim20193dsg, wald2020learning}. 
These approaches have enabled building real-time systems that can dynamically build up hierarchical 3D scene representations~\cite{rosinol2021kimera, hughes2022hydra, wu2021scenegraphfusion}, and more recently shown that various robotics planning tasks can benefit from efficiency and compactness of 3DSGs ~\cite{agia2022taskography, rana2023sayplan}.
However, existing work on building 3D scene graphs has been confined to the closed-vocabulary setting, limiting their applicability to a small set of tasks.

\subsection{Overview of Our Contribution}
In this work, we mitigate all the aforementioned limitations and propose \textbf{\coolname{}}, an open-vocabulary and object-centric 3D representation for robot perception and planning.
In \coolname, each object is represented as a node with geometric and semantic features, and relationships among objects are encoded in the graph edges. 
At the core of \coolname{} is an object-centric 3D mapping technique that integrates geometric cues from conventional 3D mapping systems, and semantic cues from vision and language foundation models~\cite{oquab2023dinov2, clip, kirillov2023sam, liu2023grounding,zhang2023ram, llava}.
Objects are assigned language tags by leveraging large language models (LLMs)~\cite{gpt4} and large vision-language models (LVLMs)~\cite{llava}, which provide semantically rich descriptions and enable free-form language querying, all while using off-the-shelf models (no training/finetuning). 
The scene graph structure allows us to efficiently represent large scenes with a low memory footprint and makes for efficient task planning.

In experiments, we demonstrate that \coolname~ is able to discover, map, and caption a large number of objects in a scene. Further, we conduct real-world trials on multiple robot platforms over a wide range of downstream tasks, including manipulation, navigation, localization, and map updates. 
To summarize, our key \textbf{contributions} are:
\begin{itemize}
  \item We propose a novel \underline{object-centric mapping system} that integrates geometric cues from traditional 3D mapping systems and semantic cues from 2D foundation models. %
  \item We construct \underline{open-vocabulary 3D scene graphs}; efficient and structured semantic abstractions for perception and planning.
  \item We implement \coolname{} on \underline{real-world} wheeled and legged robotic platforms and demonstrate a number of downstream perception and planning capabilities for complex/abstract language queries.
\end{itemize}

\section{Method}
\label{sec:approach}

\setlength{\abovedisplayskip}{0pt}
\setlength{\belowdisplayskip}{3pt}

\newcommand{\obs}{I}           %
\newcommand{\orgb}{\obs^{\text{rgb}}}  %
\newcommand{\odep}{\obs^{\text{depth}}} %
\newcommand{\opos}{\theta}  %
\newcommand{\omask}{\mathbf{m}} %
\newcommand{\mfeat}{\mathbf{f}} %
\newcommand{\objf}{\mathbf{f_o}} %
\newcommand{\mbsp}{\mathbf{p}}  %
\newcommand{\objp}{\mathbf{p_o}} %

\newcommand{\ibox}{\mathbf{b}} %
\newcommand{\pbox}{\ibox^{\text{3D}}} %

\newcommand*{\seg}{\ensuremath{\text{Seg}}} %
\newcommand{\SE}{\text{Embed}}        %
\newcommand{\lang}{\text{LVLM}}        %
\newcommand{\glang}{\text{LLM}}        %

\newcommand{\objmap}{\mathcal{M}} %
\newcommand{\obj}{\mathbf{o}}      %
\newcommand{\objset}{\mathbf{O}}   %
\newcommand{\edg}{\mathbf{e}}      %
\newcommand{\edgeset}{\mathbf{E}}  %
\newcommand{\capp}{\mathbf{c}}      %
\newcommand{\rcapp}{\hat{\capp}}      %

\newcommand{\semSim}{\phi_{\text{sem}}}  %
\newcommand{\geoSim}{\phi_{\text{geo}}}  %
\newcommand{\nnratio}{\text{nnratio}}  %
\newcommand{\overallSim}{\phi}          %
\newcommand{\simMat}{\mathbf{\phi}}     %
\newcommand{\newofeat}{\objf_{\text{new}}}  %
\newcommand{\bbodt}{\epsilon_b} %
\newcommand{\bbmat}{\mathbf{\psi}} %
\newcommand{\oldofeat}{\objf_{\text{old}}}  %
\newcommand{\numDet}{n}              %
\newcommand{\sminus}{\text{-}}  %

\newcommand{\imgseq}{\mathcal{I}} %

\newcommand{\wedgev}{w}      %
\newcommand{\ovThresh}{\alpha}  %

\newcommand{\mstree}{\mathcal{T}}  %

\begin{figure*}
    \centering
    \includegraphics[width=\textwidth, trim={0 0 1.4cm 0}]{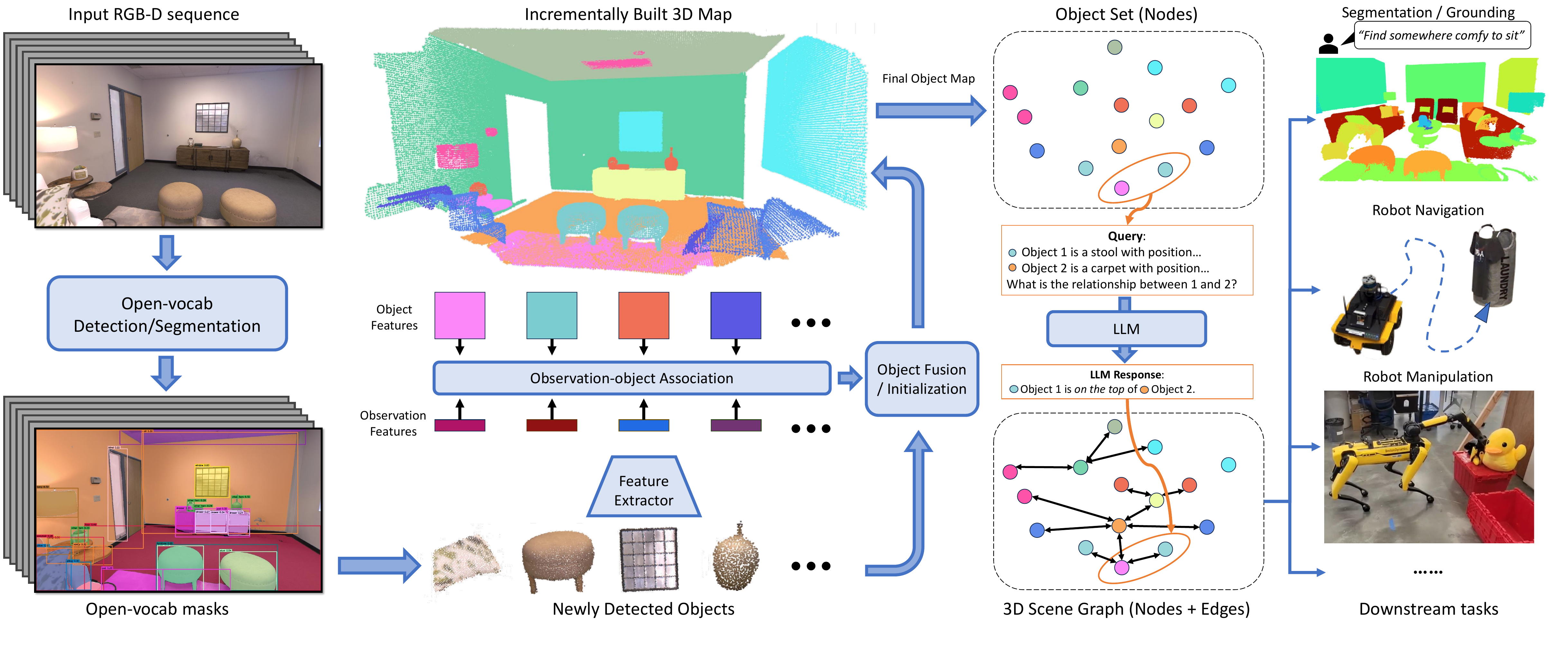}
    \vspace{-2.5em}
    \caption{\coolname{} builds an open-vocabulary 3D scene graph from a sequence of posed RGB-D images. We use generic instance segmentation models to segment regions from RGB images, extract semantic feature vectors for each, and project them to a 3D point cloud. These regions are incrementally associated and fused from multiple views, resulting in a set of 3D objects and associated vision (and language) descriptors. Then large vision and language models are used to caption each mapped 3D objects and derive inter-object relations, which generates the edges to connect the set of objects and form a graph. The resulting 3D scene graph provides a structured and comprehensive understanding of the scene and can further be easily translated to a text description, useful for LLM-based task planning.
    }
    \label{fig:pipeline}
    \vspace{-1.5em}
\end{figure*}

\coolname~builds a compact, semantically rich representation of a 3D environment. 
Given a set of posed RGB-D frames, we run a class-agnostic segmentation model to obtain candidate objects, associate them across multiple views using geometric and semantic similarity measures, and instantiate nodes in a 3D scene graph.
We then use an LVLM to caption each node and an LLM to infer relationships between adjoining nodes, which results in edges in the scene graph.
This resultant scene graph is open-vocabulary, encapsulates object properties, and can be used for a multitude of downstream tasks including segmentation, object grounding, navigation, manipulation, localization, and remapping.
The approach is illustrated in Fig.~\ref{fig:pipeline}.

\subsection{Object-based 3D Mapping}
\label{sec:objectmapping}

\textbf{Object-centric 3D representation}: Given a sequence of RGB-D observations $\imgseq = \{ \obs_1, \obs_2, \ldots, \obs_t \}$, \coolname~constructs a map, a 3D scene graph, $\objmap_t = \langle \objset_t, \edgeset_t \rangle$, where $\objset_t = \{\obj_{j}\}_{j=1...J}$ and $\edgeset_t = \{\edg_{k}\}_{k=1...K}$ represent the sets of objects and edges, respectively. Each object $\obj_{j}$ is characterized by a 3D point cloud $\objp_j$ and a semantic feature vector $\objf_j$. 
This map is built incrementally, incorporating each incoming frame $\obs_t = \langle \orgb_t, \odep_t, \opos_t \rangle$ (color image, depth image, pose) into the existing object set $\objset_{t-1}$, 
by either adding to existing objects or instantiating new ones.

\textbf{Class-agnostic 2D Segmentation}:
When processing frame $\obs_t$, a class-agnostic segmentation model $\seg(\cdot)$ is used 
to obtain a set of masks $\{\omask_{t,i}\}_{i=1...M} = \seg(\orgb_t)$ corresponding to candidate objects\footnote{Without loss of generality, $\seg(\cdot)$ may be replaced by open-/closed-vocabulary models to build category-specific mapping systems.}.
Each extracted mask $\omask_{t,i}$ is then passed to a visual feature extractor (CLIP~\cite{clip}, DINO~\cite{oquab2023dinov2}) to obtain a visual descriptor $\mfeat_{t,i}=\SE(\orgb_t, \omask_{t,i})$.
Each masked region is projected to 3D, denoised using DBSCAN clustering, and transformed to the map frame.
This results in a pointcloud $\mbsp_{t,i}$
and its corresponding unit-normalized semantic feature vector $\mfeat_{t,i}$.

\textbf{Object Association}: For every newly detected object $\langle \mbsp_{t,i}, \mfeat_{t,i} \rangle$, we compute semantic and geometric similarity with respect to all objects $\obj_{t\sminus1,j}=\langle \objp_j, \objf_j \rangle$ in the map that shares any partial geometric overlap.
The geometric similarity $\geoSim(i, j) =  \nnratio (\mbsp_{t,i} , \objp_{j} )$ is the proportion of points in point cloud $\mbsp_{t,i}$ that have nearest neighbors in point cloud $\objp_{j}$, within a distance threshold of $\delta_{\text{nn}}$.
The semantic similarity $\semSim(i, j) = \mfeat_{t,i}^T \objf_{j} /2+1/2$ is the normalized cosine distance between the corresponding visual descriptors.\footnote{For the sake of brevity, we only describe the best-performing geometric and semantic similarity measures. For an exhaustive list of alternatives, please see our project website and code.}
The overall similarity measure $\overallSim(i, j)$ is a sum of both: $ \overallSim(i, j) = \semSim(i, j) + \geoSim(i, j) $.
We perform object association by a greedy assignment\footnote{While we also experimented with optimal assignment strategies such as the Hungarian algorithm, we experimentally determined them to be slower and offer only a minuscule improvement over greedy association.} strategy where each new detection is matched with an existing object with the highest similarity score.
If no match is found with a similarity higher than $\delta_{\text{sim}}$, we initialize a new object.

\textbf{Object Fusion}: If a detection $\obj_{t\sminus1,j}$ is associated with a mapped object $\obj_j$, we fuse the detection with the map. 
This is achieved by updating the object semantic feature as $\objf_j = (\numDet_{\obj_j} \objf_j + \mfeat_{t,i}) / (\numDet_{\obj_j}+1)$, where $\numDet_{\obj_j}$ is the number of detections that have been associated to $\obj_j$ so far;
and updating the pointcloud as $\mbsp_{t,i} \cup \objp_{j}$, followed by down-sampling to remove redundant points.

\textbf{Node Captioning}: 
Once the entire image sequence has been processed, a vision-language model, denoted $\lang ( \cdot )$, is used to generate object captions. For each object, the associated image crops from the \emph{best}\footnote{We maintain a running index of the number of noise-free points each view contributes to the object point cloud.} 10 views are passed to the language model with the prompt ``\textit{describe the central object in the image}'' to generate a set of initial rough captions $\rcapp_j = \{\rcapp_{j,1}, \rcapp_{j,2}, \dots , \rcapp_{j,10} \} $ for each detected object $\obj_j$. Each set of captions is then refined to the final caption by passing $\rcapp_j$ to another language model $\glang (\cdot)$ with a prompt instruction to summarize the initial captions into a coherent and accurate final caption $\capp_j$.

\relax{

}

\subsection{Scene Graph Generation}

Given the set of 3D objects $\objset_T$ obtained from the previous step, we estimate their spatial relationships, i.e., the edges $\edgeset_T$, to complete the 3D scene graph. 
We do this by first estimating potential connectivity among object nodes based on their spatial overlaps. We compute the 3D bounding box IoU between every pair of object nodes to obtain a similarity matrix (i.e., a dense graph), which we prune by estimating a minimum spanning tree (MST), resulting
in a refined set of potential edges among the objects. 
To further determine the semantic relationships, for each edge in the MST, we input the information about the object pair, consisting of object captions and 3D location, to a language model $\glang$. The prompt instructs the model to describe the likely spatial relationship between the objects, such as ``\textit{a on b}'' or ``\textit{b in a}'', along with the underlying reasoning. 
The model outputs a relationship label with an explanation detailing the rationale. 
The use of an LLM allows us to extend the nominal edge type defined above to other output relationships a language model can interpret, such as
``a backpack \emph{may be stored in} a closet'' and ``sheets of paper \emph{may be recycled in} a trash can''. 
This results in an open-vocabulary 3D scene graph $\objmap_T = (\objset_T, \edgeset_T)$, a compact and efficient representation for use in downstream tasks.

\subsection{Robotic Task Planning through LLMs}

To enable users to carry out tasks described in natural language queries, we interface the scene graph $\objmap_T$ with an LLM.
For each object in $\objset_T$, we construct JSON-structured text descriptions that include information about its 3D location (bounding box) and its node caption.
Given a text query, we task the LLM to identify the most relevant object in the scene.
We then pass the 3D pose of this object to the appropriate pipeline for the downstream task (e.g., grasping, navigation). This integration of \coolname{} with an LLM is easy to implement, and enables a wide range of open-vocabulary tasks by giving robots access to the semantic properties of surrounding objects\footnote{For large scenes where the description length of the scene graph exceeds the context length of the LLM, one can easily substitute in alternative (concurrent) LLM planners~\cite{rana2023sayplan}.} (see Sec.~\ref{sec:results}).

\subsection{Implementation Details}

The modularity of \coolname{} enables any appropriate open/closed-vocabulary segmentation model, LLM, or LVLM to be employed.
Our experiments use Segment-Anything (SAM)~\cite{kirillov2023sam} as the segmentation model $\seg(\cdot)$, and the CLIP image encoder~\cite{clip} as the feature extractor $\SE(\cdot)$. We use LLaVA~\cite{llava} as the vision-language model $\lang$ and GPT-4~\cite{gpt4} (\texttt{gpt-4-0613}) for our $\glang$. The voxel size for point cloud downsampling and nearest neighbor threshold $\delta_{\text{nn}}$ are both $2.5$cm. We use $1.1$ for the association threshold $\delta_{\text{sim}}$. 

We also develop a variant of our system, \coolnamedetect~(\coolabbrdetect), where we employ an image tagging model (RAM~\cite{zhang2023ram}) to list the object classes present in the image and an open-vocabulary 2D detector (Grounding DINO~\cite{liu2023grounding}) to obtain object bounding boxes\footnote{We discard the (often noisy) \emph{tags} produced by the image tagging model, relying instead on our node captions.}.
In this variant, we need to separately handle detected background objects (wall, ceiling, floor) by
merging them regardless of their similarity scores.

\section{Experiments}
\label{sec:results}

\subsection{Scene Graph Construction}
\label{sec:sg_construct}

We first evaluate the accuracy of the 3D scene graphs output by the \coolname{} system in Table~\ref{tab:sg-accuracy}. For each scene in the Replica dataset~\cite{replica}, we report scene graph accuracy metrics for both \emph{CG} and the detector-variant \emph{CG-D}.
The open-vocabulary nature of our system makes automated evaluation of the quality of nodes and edges in the scene graph challenging.
We instead evaluate the scene graph by engaging human evaluators on Amazon Mechanical Turk (AMT).
For each node, we compute precision as the fraction of nodes for which at least 2 of 3 human evaluators deem the node caption correct.
We also report the number of valid objects retrieved by each variant by asking evaluators whether they deem each node a valid object.
Both CG and CG-D identify a number of valid objects in each scene, and incur only a small number (0-5) of duplicate detections.
The node labels are accurate about $70\%$ of the time; most of the errors are incurred due to errors made by the LVLM employed (LLaVA~\cite{llava}).
The edges (spatial relationships) are labeled with a high degree of accuracy ($90\%$ on average).

\begin{table}[!t]
    \adjustbox{max width=\linewidth}{%
    \centering
    \begin{tabular}{lccccc}
        \toprule
        & scene & node prec. & valid objects & duplicates & edge prec. \\
        \midrule
        \multirow{8}{*}{CG}
        & room0 & 0.78 & 54 & 3 & 0.91 \\
        & room1 & 0.77 & 43 & 4 & 0.93 \\
        & room2 & 0.66 & 47 & 4 & 1.0 \\
        & office0 & 0.65 & 44 & 2 & 0.88  \\
        & office1 & 0.65 & 23 & 0 & 0.9 \\
        & office2 & 0.75 & 44 & 3 & 0.82 \\
        & office3 & 0.68 & 60 & 5 & 0.79 \\
        & \cellcolor{gray!25} \textbf{Average} & \cellcolor{gray!25} \textbf{0.71} & \cellcolor{gray!25} - & \cellcolor{gray!25} - & \cellcolor{gray!25} \textbf{0.88} \\
        \midrule
        \multirow{8}{*}{CG-D}
        & room0 & 0.56 & 60 & 4 & 0.87 \\
        & room1 & 0.70 & 40 & 3 & 0.93 \\
        & room2 & 0.54 & 49 & 2 & 0.93 \\
        & office0 & 0.59 & 35 & 0 & 1.0 \\
        & office1 & 0.49 & 24 & 2 & 0.9 \\
        & office2 & 0.67 & 47 & 3 & 0.88 \\
        & office3 & 0.71 & 59 & 1 & 0.83 \\
        & \cellcolor{gray!25} \textbf{Average} & \cellcolor{gray!25} \textbf{0.61} & \cellcolor{gray!25} - & \cellcolor{gray!25} - & \cellcolor{gray!25} \textbf{0.91} \\ 
        \bottomrule
    \end{tabular}
    } %
    \caption{\textbf{Accuracy of constructed scene graphs}: node precision indicates the accuracy of the label for each node (as measured by a human evaluator); valid objects is the number of human-recognizable objects (mturkers used) discovered by our system; duplicates are the number of redundant detections; edge precision indicates the accuracy of each estimated spatial relationship (again, as evaluated by an mturker)}
    \label{tab:sg-accuracy}
    \vspace{-2em}
\end{table}

\subsection{3D Semantic Segmentation}

\coolname~focuses on the construction of the open-vocabulary 3D scene graphs for scene understanding and planning. For completeness, in this section, we also use an open-vocabulary 3D semantic segmentation task to evaluate the quality of the obtained 3D maps. 
To generate the semantic segmentation, given a set of class names, we compute the similarity between the fused semantic feature of each object node and the CLIP text embeddings of the phrase \verb|an image of {class}|. Then the points associated with each object are assigned to the class with the highest similarity, which gives a point cloud with dense class labels. 
In Table~\ref{tab:semseg}, we report the semantic segmentation results on the Replica~\cite{replica} dataset, following the evaluation protocol used in ConceptFusion~\cite{conceptfusion}. We also provide an additional baseline, ConceptFusion+SAM, by replacing the Mask2Former used in ConceptFusion with the more performant SAM~\cite{kirillov2023sam} model. As shown in Table~\ref{tab:semseg}, the proposed \coolname~performs comparably with or better than ConceptFusion, 
which has a much larger memory footprint. 

\begin{table}[!t]
    \centering
    \begin{tabular}{llrr}\toprule
                            & Method                         & mAcc                & F-mIoU \\ \midrule
\multirow{3}{*}{Privileged} & CLIPSeg (rd64-uni)~\cite{clipseg}        & 28.21              & 39.84  \\
                            & LSeg~\cite{lseg}                           & 33.39                     & 51.54                      \\
                            & OpenSeg~\cite{openseg}         & 41.19                     & 53.74                      \\ \midrule
\multirow{6}{*}{Zero-shot}  & MaskCLIP~\cite{maskclip}       & 4.53                      & 0.94                       \\
                            & Mask2former + Global CLIP feat & 10.42                     & 13.11                      \\
                            & ConceptFusion~\cite{conceptfusion}  & 24.16                & 31.31                      \\
                            & ConceptFusion~\cite{conceptfusion} + SAM~\cite{kirillov2023sam}  & 31.53                     & \textbf{38.70}                      \\
                            & \coolname~(Ours)                &  \textbf{40.63}       &   35.95                    \\
                            & \coolnamedetect~(Ours)        &   38.72                 &   35.82                   \\ \bottomrule
\end{tabular}
    \caption{Open-vocabulary semantic segmentation on the Replica~\cite{replica} dataset. \textbf{Privileged} methods specifically finetune the pretrained models for semantic segmentation. \textbf{Zero-shot} approaches do not need any finetuning and are evaluated off the shelf. }
    \label{tab:semseg}
    \vspace{-2em}
\end{table}

\subsection{Object Retrieval based on Text Queries}

We assess the capability of \coolname~to handle complex semantic queries, focusing on three key types.
\begin{itemize}
    \item Descriptive: E.g., \emph{A potted plant}.
    \item Affordance: E.g., \emph{Something to use for temporarily securing a broken zipper}.
    \item Negation: E.g., \emph{Something to drink other than soda}.
\end{itemize}

We evaluate on the Replica dataset~\cite{replica} and a real-world scan of the \href{https://montrealrobotics.ca/}{REAL Lab}, where we staged a number of items including clothes, tools, and toys. 
For Replica, human evaluators on AMT annotate captions for SAM mask proposals, which serve as both ground truth labels and descriptive queries.
We created 5 affordance and negation queries for each scene type (office \& room) in Replica and 10 queries of each type for the lab scan, ensuring that each query corresponds to at least one relevant object. We manually select relevant objects as ground truth for each query. 

We use two object retrieval strategies: CLIP-based and LLM-based. CLIP selects the object with the highest similarity to the query's embedding, while the LLM goes through the scenegraph nodes to identify the object with the most relevant caption.
Table~\ref{tab:objret} shows that CLIP excels with descriptive queries but struggles with complex affordance and negation queries \cite{du2020compositional}. For example, CLIP inaccurately retrieves a backpack for the broken zipper query, whereas the LLM correctly identifies a roll of tape. The LLM performs well across the board, but is limited by the accuracy of the node captions, as discussed in Section~\ref{sec:sg_construct}. Since the lab has a larger variety of objects to choose from, the LLM finds compatible objects for complex queries more reliably there.

\subsection{Complex Visual-Language Queries}

To assess the performance of~\coolname~in a real-world environment, we carry out navigation experiments in the \href{https://montrealrobotics.ca/}{REAL Lab} scene with a Clearpath Jackal UGV. The robot is equipped with a VLP-16 LiDAR and a forward-facing Realsense D435i camera. 

The Jackal needs to respond to abstract user queries and navigate to the most relevant object (Figure~\ref{fig:splash}). By using an LVLM~\cite{llava} to add a description of the current camera image to the text prompt, the robot can also answer visual queries. For example, when shown a picture of Michael Jordan and prompted with \texttt{Something this guy would play with}, the robot finds a basketball.

\subsection{Object Search and Traversability Estimation}

In this section, we showcase how the interaction between the~\coolname~representation and an LLM can enable a mobile robot to access a vast knowledge base of everyday objects. Specifically, we prompt an LLM to infer two additional object properties from~\coolname~captions: i)~the location where a given object is typically found,  and ii)~if the object can be safely pushed or traversed by the Jackal robot. We design two tasks around the LLM predictions.

\begin{figure*}
    \centering
    \includegraphics[width=0.85\linewidth]{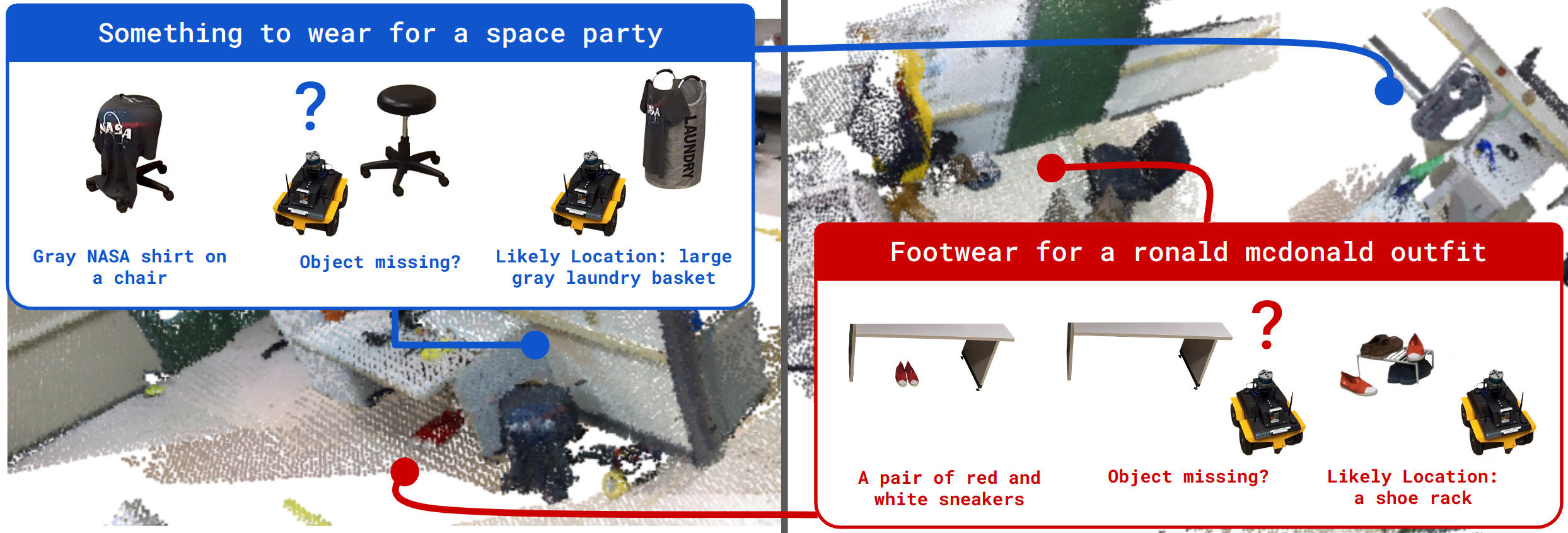}
    \caption{A Jackal robot answering user queries using the~\coolname~representation of a lab environment. We first query an LLM to identify the most relevant object given the user query, then validate with an LVLM if the target object if is at the expected location. If not, we query the LLM again to find a likely location or container for the missing object. (Blue)~When prompted with \texttt{something to wear for a space party}, the Jackal attempts to find a grey shirt with a NASA logo. After failing to detect the shirt at the expected location, the LLM reasons that it could likely be in the laundry bag. (Red)~The Jackal searches for red and white sneakers after receiving the user query ~\texttt{footwear for a Ronald McDonald outfit}. The LLM redirects the robot to a shoe rack after failing to detect the sneakers where they initially appeared on the map.}
    \vspace{-1em}
    \label{fig:jackal_queries}
\end{figure*}

\textbf{Object search}: The robot receives an abstract user query and must navigate to the most relevant object in the~\coolname~map. Using an  LVLM~\cite{llava}, the robot then checks if the object is at the expected location. If not, it queries an LLM to find a new plausible location given the captions of the other objects in the representation. In our prompt, we nudge the LLM to consider typical containers or storage locations. We illustrate two such queries where the target object is moved in Figure~\ref{fig:jackal_queries}.

\begin{table}[!t]
    \centering
\begin{tabular}{lllrrrrc}
    \toprule
    Dataset & Query Type  & Model & R@1 & R@2 & R@3 & \# Queries \\
    \midrule
    \multirow{6}{*}{Replica} & \multirow{2}{*}{Descriptive} & \cellcolor{gray!25} CLIP & \cellcolor{gray!25} 0.59 & \cellcolor{gray!25} 0.82 & \cellcolor{gray!25} 0.86 & \multirow{2}{*}{20} \\
                             &                              & \cellcolor{gray!25} LLM  & \cellcolor{gray!25} 0.61 & \cellcolor{gray!25} 0.64 & \cellcolor{gray!25} 0.64 & \\
                             & \multirow{2}{*}{Affordance}  & \cellcolor{gray!50} CLIP & \cellcolor{gray!50} 0.43 & \cellcolor{gray!50} 0.57 & \cellcolor{gray!50} 0.63 & \multirow{2}{*}{5}  \\
                             &                              & \cellcolor{gray!50} LLM  & \cellcolor{gray!50} 0.57 & \cellcolor{gray!50} 0.63 & \cellcolor{gray!50} 0.66 & \\
                             & \multirow{2}{*}{Negation}    & \cellcolor{gray!25} CLIP & \cellcolor{gray!25} 0.26 & \cellcolor{gray!25} 0.60 & \cellcolor{gray!25} 0.71 & \multirow{2}{*}{5}  \\
                             &                              & \cellcolor{gray!25} LLM  & \cellcolor{gray!25} 0.80 & \cellcolor{gray!25} 0.89 & \cellcolor{gray!25} 0.97 & \\
    \midrule
    \multirow{6}{*}{Lab}     & \multirow{2}{*}{Descriptive} & \cellcolor{gray!25} CLIP & \cellcolor{gray!25} 1.00 & \cellcolor{gray!25} --   & \cellcolor{gray!25} --   & \multirow{2}{*}{10} \\
                             &                              & \cellcolor{gray!25} LLM  & \cellcolor{gray!25} 1.00 & \cellcolor{gray!25} --   & \cellcolor{gray!25} --   & \\
                             & \multirow{2}{*}{Affordance}  & \cellcolor{gray!50} CLIP & \cellcolor{gray!50} 0.40 & \cellcolor{gray!50} 0.60 & \cellcolor{gray!50} 0.60 & \multirow{2}{*}{10} \\
                             &                              & \cellcolor{gray!50} LLM  & \cellcolor{gray!50} 1.00 & \cellcolor{gray!50} -- & \cellcolor{gray!50} -- & \\
                             & \multirow{2}{*}{Negation}    & \cellcolor{gray!25} CLIP & \cellcolor{gray!25} 0.00 & \cellcolor{gray!25} --   & \cellcolor{gray!25} --   & \multirow{2}{*}{10} \\
                             &                              & \cellcolor{gray!25} LLM  & \cellcolor{gray!25} 1.00 & \cellcolor{gray!25} --   & \cellcolor{gray!25} --   & \\
    \bottomrule
\end{tabular}

    \caption{Object retrieval from text queries on the Replica and REAL Lab scenes. We measure the top-1, top-2, and top-3 recall. CLIP refers to object retrieval using cosine similarity, whereas LLM refers to having an LLM parse the scene graph and return the most relevant object.
    }

    \label{tab:objret}
    \vspace{-2em}
\end{table}

\begin{figure}
    \centering
    \includegraphics[width=\linewidth]{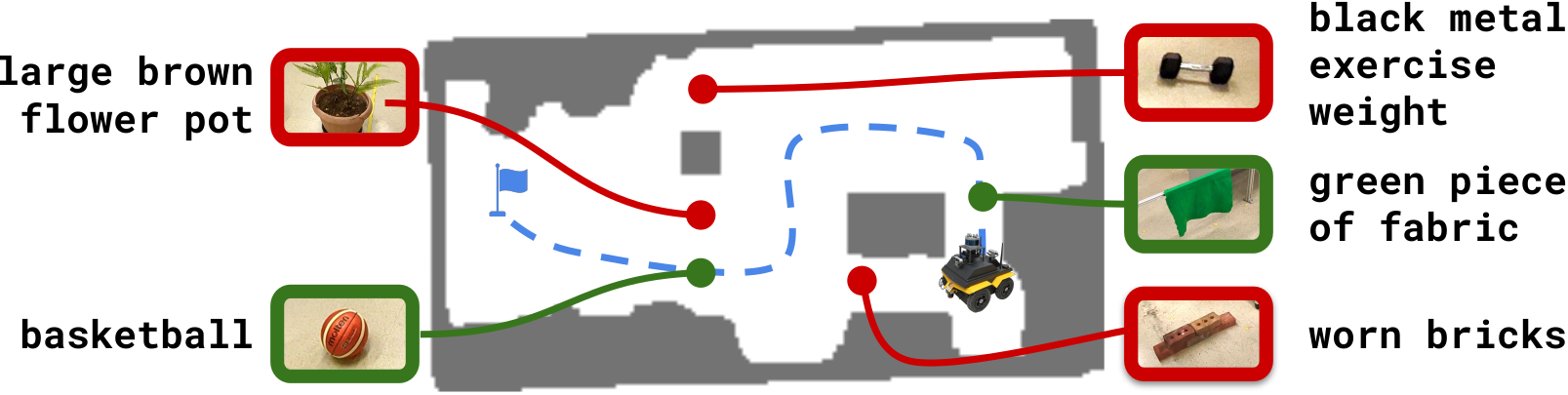}
    \caption{The Jackal robot solving a traversability challenge. All paths to the goal are obstructed by objects. We query an LLM to identify which objects can be safely pushed or traversed by the robot (green) and which objects would be too heavy or hinder the robot's movement (red). The LLM relies on the~\coolname~node captions to make traversability predictions and we add the non-traversable objects to the Jackal costmap for path planning. The Jackal successfully reaches the goal by going through a curtain and pushing a basketball, while also avoiding contact with bricks, an iron dumbbell, and a flower pot.}
    \label{fig:traversability}
    \vspace{-1.5em}
\end{figure}

\textbf{Traversability estimation}: As shown in Fig.~\ref{fig:traversability}, we design a real-world scenario where the robot finds itself enclaved by objects. In this scenario, the robot must push around multiple objects and create a path to the goal state.
While traversability can be learned through experience~\cite{levine2023learning}, we show that grounding LLM knowledge in a 3D map can grant similar capabilities to robotic agents.

\subsection{Open-Vocabulary Pick and Place}

To illustrate how~\coolname~can act as the perception backbone for open-vocabulary mobile manipulation, we conducted a series of experiments with a Boston Dynamics Spot Arm robot. Using an onboard RGBD camera and a~\coolname~representation of the scene, the Spot robot responds to the query \texttt{cuddly quacker} by grabbing a duck plush toy and placing it in a nearby box~(Figure~\ref{fig:splash}). In the supplementary video, Spot completes a similar grasping maneuver with a mango when prompted with the query \texttt{something healthy to eat}.

\subsection{Localization and Map Updates}

\coolname~can also be used for object-based localization and map updates. We showcase this with a 3-DoF ($x$, $y$ and yaw) localization and remapping task in the AI2Thor~\cite{ai2thor, procthor} simulation environment, where a mobile robot uses a particle filter to localize in a pre-built \coolname~map of the environment. During the observation update step of particle filtering, the robot's detections are matched against the objects in the map based on the hypothesized pose, in a similar way as described in Section~\ref{sec:objectmapping}. The matching results are aggregated into a single observation score for weighting the pose hypothesis.
During this process, previously observed objects are removed if they are not observed by the robot and new objects can also be added. We provide a demonstration of this localization and map updating approach in the supplementary video material.

\subsection{Limitations}
\label{sec:limitations}

Despite its impressive performance, \coolname{} has failure modes that remain to be addressed in future work. First, node captioning incurs errors due to the current limitations of LVLMs like LLaVA~\cite{llava}. Second, our 3D scene graph occasionally misses small or thin objects and makes duplicate detections. This impacts downstream planning, particularly when the incorrect detection is crucial to planning success. Additionally, the computational and economic costs of our system include multiple LVLM (LLaVA~\cite{llava}) and one or more proprietary LLM inference(s) when building and querying the scenegraph, which may be significant.

\section{Concurrent Work}
\label{sec:related_work}

We briefly review recent and unpublished pre-prints that are exploring themes related to open-vocabulary object-based factorization of 3D scenes. 
Concurrently to us, \cite{takmaz2023openmask3d, lu2023ovir3d} have explored open-vocabulary object-based factorization of 3D scenes.
Where~\cite{takmaz2023openmask3d} assumes a pre-built point cloud map of the scene, \cite{lu2023ovir3d} builds a map on the go.
Both approaches associate CLIP descriptors to the reconstruction, resulting in performance comparable to our system's CLIP variant, which struggles with queries involving complex affordances and negation, as shown in Table \ref{tab:objret}.
OGSV~\cite{chang2023ogsv} is closer to our setting, building an open-vocabulary 3D scene graph from RGB-D images.
However,~\cite{chang2023ogsv} relies on a (closed-set) graph neural network to predict object relationships; whereas \coolname{} relies on the capabilities of modern LLMs, eliminating the need to train an object relation model.

\section{Conclusion}

In this paper, we introduced \coolname, a novel approach to open-vocab object-centric 3D scene representation that addresses key limitations in the existing landscape of dense and implicit representations. Through effective integration of foundational 2D models, \coolname~significantly mitigates memory constraints, provides relational information among objects, and allows for dynamic updates to the scene—three pervasive challenges in current methods. 
Experimental evidence underscores \coolname' robustness and extensibility, highlighting its superiority over existing baselines for a variety of real-world tasks including manipulation and navigation.
The versatility of our framework also accommodates a broad range of downstream applications, thereby opening new avenues for innovation in robot perception and planning. Future work may delve into integrating temporal dynamics into the model and assessing its performance in less structured, more challenging environments.

\section*{Acknowledgments}

This project was supported in part (KM, AT, JBT) by grants from the Army Research Laboratory (grant W911NF1820218), office of naval research (MURI). FS and LP thank NSERC for funding support. LP acknowledges support from the Canada CIFAR AI Chairs program.
All findings, opinions, and conclusions expressed in this manuscript solely reflect the views of the authors; and not that of our research sponsors.

\section*{Appendix}

\setcounter{section}{0}
\setcounter{equation}{0}
\setcounter{figure}{0}
\setcounter{table}{0}

\renewcommand{\thesection}{A\arabic{section}}
\renewcommand{\thefigure}{A.\arabic{figure}}
\renewcommand{\thetable}{A.\arabic{table}}
\renewcommand{\theequation}{A.\arabic{equation}}

\section{Contribution Statement}

\textbf{Qiao Gu, Ali Kuwajerwala, Sacha Morin}, and \textbf{Krishna Murthy} were instrumental in the development, integration, and deployment of \coolname{}. These authors were responsible for writing the majority of the manuscript.

\textbf{Qiao} spearheaded the implementation of the object-based mapping, localization and map update system, prototyped the object captioning module, and conducted the segmentation experiments. 

\textbf{Ali} wrote initial prototypes of the mapping pipeline, coordinated the real-robot experiments at the REAL Lab, implemented the integration with the LLM planners and object retrieval experiments, and performed a significant amount of the hardware setup. 

\textbf{Sacha} was at the forefront of deploying the system on robot navigation, object search, and led the traversability experiment, implementing a variety of crucial robot-side functionalities. 

\textbf{Krishna} crafted initial prototypes of the detector-free mapping system, scene graph construction, and vision/language model interfaces, also coordinating the project. 

\textbf{Bipasha Sen} and \textbf{Aditya Agarwal} integrated \coolname{} with a robotic manipulation platform for open-vocabulary pick and place demonstrations, with \textbf{Kirsty Ellis} assisting them in setting up the manipulation platform.

\textbf{Corban Rivera} and \textbf{William Paul} deployed \coolname{} on-board a Spot mini robot, showcasing the mobile manipulation capabilities enabled by our system.

\textbf{Chuang Gan} helped qualitatively evaluate our system against end-to-end learned approaches like 3D-LLM.

\textbf{Rama Chellappa} and \textbf{Celso de Melo} provided adivce on real-world demonstrations involving mobile manipulation.

\textbf{Josh Tenenbaum} and \textbf{Antonio Torralba} contributed cognitive science and computer vision perspectives respectively, which shaped the experiments evaluating scene-graph construction accuracy.

\textbf{Florian Shkurti} and \textbf{Liam Paull} provided equal advisory support on this project, contributing to brainstorming and critical review processes throughout, and writing/proofreading significant sections of the paper.

\section{3D Scene Graph: Generating Node Captions}

\definecolor{codeblue}{rgb}{0.38039216, 0.61568627, .8       }
\definecolor{codepink}{rgb}{1.        , 0.41568627, 0.83529412}
\definecolor{codegray}{rgb}{0.5,0.5,0.5}
\definecolor{codepurple}{rgb}{0.58,0,0.82}
\definecolor{backcolour}{rgb}{0.98,0.98,0.98}

\lstdefinestyle{mystyle}{
    backgroundcolor=\color{backcolour},   
    commentstyle=\color{codeblue},
    keywordstyle=\color{codepink},
    numberstyle=\tiny\color{codegray},
    stringstyle=\color{codepurple},
    basicstyle=\ttfamily\scriptsize,
    breakatwhitespace=false,         
    breaklines=true,                 
    captionpos=b,                    
    keepspaces=true,                 
    showspaces=false,                
    showstringspaces=false,
    showtabs=false,                  
    tabsize=2
}
\lstset{style=mystyle}

\lstdefinelanguage{gptprompt}{
  basicstyle=\ttfamily\footnotesize,
  keywords = {id, bbox\_extent, bbox\_center, object\_tag, caption, inferred\_query, relevant\_objects, query\_achievable, final\_relevant\_objects, explanation, query\_text},
  keywordstyle=\color{red},
  keywords = [2]{User, LLM-Planner},
  keywordstyle = [2]\color{blue},
}

Once we build an object-level map of the scene using the methodology described in Sec.~\ref{sec:objectmapping}, we extract and summarize captions for each object.
We first extract upto the $10$ \emph{most-informative} views for each object, by tracking the number of (noise-free) 3D points that each image segment contributes to an object in the map\footnote{We track these statistics throughout the mapping lifecycle; meaning that we do not impose any additional computational overhead to determine the $10$ best views per object}. Intuitively, these views offer the best image views for the object.
We run each view through an LVLM, here LLaVA-7B~\cite{llava}, to generate an image caption. We use the same prompt across all images: \emph{describe the central object in this image}.

We found the captions generated by LLaVA-7B to be incoherent or unreliable across all viewpoints. To alleviate this, we employed GPT-4 as a caption summarizer, to map all of the LLaVA-7B captions to a coherent object tag (or optionally, declare the object as an invalid detection).
We use the following GPT-4 system prompt:

\begin{lstlisting}[language=gptprompt, caption=GPT-4 system prompt used for caption summarization, escapeinside={(*@}{@*)}]
(*@
\begin{minipage}{\linewidth}
Identify and describe objects in scenes. Input and output must be in JSON format. The input field 'captions' contains a list of image captions aiming to identify objects. Output 'summary' as a concise description of the identified object(s). An object mentioned multiple times is likely accurate. If various objects are repeated and a container/surface is noted such as a shelf or table, assume the (repeated) objects are on that container/surface. For unrelated, non-repeating (or empty) captions, summarize as 'conflicting (or empty) captions about [objects]' and set 'object\_tag' to 'invalid'. Output 'possible\_tags' listing potential object categories. Set 'object\_tag' as the conclusive identification. Focus on indoor object types, as the input captions are from indoor scans.
\end{minipage}
@*)
\end{lstlisting}

\section{LLM Planner: Implementation details}
\label{app:sec:llm-planner-details}

For task planning over 3D scene graphs, we use GPT-4 (\texttt{gpt-4-0613}) with a context length of 8K tokens\footnote{We also prototyped variants of this approach on off-the-shelf LLMs with larger context lengths, such as Claude-2 with a context length of 32K tokens, and found it to work reliably.}. We first convert each node in the 3D scene graph into a structured text format (here, a JSON string). Each entry in the JSON list corresponds to one object in the scene, and contains the following attributes:
\begin{enumerate}
    \item \textit{object id}: a unique (numerical) object identifier
    \item \textit{bounding box extents}: dimensions of each side of the bounding cuboid
    \item \textit{bounding box center}: centroid of the object bounding cuboid
    \item \textit{object tag}: a brief tag describing the object
    \item \textit{caption}: a one-sentence caption (possibly encoding mode details than present in the object tag
\end{enumerate}

Here is a sample snippet from the scene graph for the \texttt{room0} scene of the Replica~\cite{replica} dataset.

\begin{lstlisting}[language=gptprompt, caption=Sample text entries in the 3D scene graph]
[
{
    id: 2,
    bbox_extent: [2.0, 0.7, 0.6],
    bbox_center: [-0.6, 1.1, -1.2],
    object_tag: wooden dresser or chest of drawers,
    caption: A wooden dresser or chest of drawers
},
{
    id: 3,
    bbox_extent: [0.6, 0.5, 0.4],
    bbox_center: [2.8, -0.4, -0.8],
    object_tag: vase,
    caption: a white, floral-patterned vase (or possibly a ceramic bowl)
},
...
...
{
    id: 110,
    bbox_extent: [1.2, 0.6, 0.0],
    bbox_center: [2.2, 2.1, 1.2],
    object_tag: light fixture,
    caption: a light fixture hanging from the ceiling
}
]
\end{lstlisting}

For planning using the above text description of the scene, we instantiate the system prompt for GPT-4 as follows:

\begin{lstlisting}[language=gptprompt, caption=System prompt for the LLM Planner interface]
The input to the model is a 3D scene described in a JSON format. Each entry in the JSON describes one object in the scene, with the following five fields:
1. "id": a unique object id
2. "bbox_extent": extents of the 3D bounding box for the object
3. "bbox_center": centroid of the 3D bounding box for the object
4. "object_tag": a brief (but sometimes inaccurate) tag categorizing the object
5. "caption": a brief caption for the object

Once you have parsed the JSON and are ready to answer questions about the scene, say "I'm ready".

The user will then begin to ask questions, and the task is to answer various user queries about the 3D scene. For each user question, respond with a JSON dictionary with the following fields:
1. "inferred_query": your interpretaion of the user query in a succinct form
2. "relevant_objects": list of relevant object ids for the user query (if applicable)
3. "query_achievable": whether or not the user-specified query is achievable using the objects and descriptions provided in the 3D scene.
4. "final_relevant_objects": A final list of objects relevant to the user-specified task. As much as possible, sort all objects in this list such that the most relevant object is listed first, followed by the second most relevant, and so on.
5. "explanation": A brief explanation of what the most relevant object(s) is(are), and how they achieve the user-specified task.
\end{lstlisting}

Here are some example GPT-4 responses on our queries

\begin{lstlisting}[language=gptprompt, caption=Sample queries and raw outputs from the LLM Planner]
User: I need a quick gift. Help!

LLM-Planner: {
inferred_query: Find suitable object for a gift.,
relevant_objects: [0, 6, 7, 23, 25, 31],
query_achievable: true,
final_relevant_objects: [6],
explanation: The most suitable object for a gift could be the ceramic vase (id 6). Vases are commonly gifted items and this one could potentially be filled with a plant or flower arrangement, making a lovely present.
}

User: Need something to use as a paperweight

LLM-Planner: {
inferred_query: Find object to use as a paperweight.,
relevant_objects: [0, 6, 7, 8, 15, 23, 29, 31],
query_achievable: true,
final_relevant_objects: [6],
explanation: The ceramic vase (id 6) could potentially be used as a paperweight due to its small size and assumed weight.
}
\end{lstlisting}

\section{Example Text Queries}
\label{app:sec:query-details}

Here we provide some of the text queries used in our experiments.

Replica Dataset Descriptive Queries:
For each room, we randomly selected 20 ground truth annotations collected via Amazon Mechanical Turk (AMT). Here is a sample from \texttt{room0} and \texttt{office0}.

\noindent \textbf{\texttt{office0} Descriptive Queries}:
\begin{enumerate}
    \item \textit{This is a trash can against the wall next to a sofa.}
    \item \textit{A chaise lounge right next to a small table.}
    \item \textit{This is a television.}
    \item \textit{This is a dropped, tiled ceiling in what appears to be a classroom for children.}
    \item \textit{This is a plant and it is next to the screens.}
    \item \textit{This is the back of a chair in front of a screen.}
    \item \textit{A small table in front of a large gray sectional couch.}
    \item \textit{This is an armless chair and it's opposite a coffee table by the sofa.}
    \item \textit{This is a plug-in and it is on the floor.}
    \item \textit{These are table legs and they are underneath the table.}
    \item \textit{These are chairs and they are next to a table.}
    \item \textit{A diner style table in front of two chairs.}
    \item \textit{These are rocks and they are on the wall.}
    \item \textit{This is the right panel of a lighted display screen.}
    \item \textit{This is a planet and it is on the wall.}
    \item \textit{This is an electronic display screen showing a map, on the wall.}
    \item \textit{This is a couch and it is between a table and the wall.}
    \item \textit{This is a garbage can and it is in front of the wall.}
    \item \textit{This is a rug and it is on the floor.}
    \item \textit{This is a table that is above the floor.}
\end{enumerate}

\noindent \textbf{\texttt{room0} Descriptive Queries}:
\begin{enumerate}
    \item \textit{This is a pillow and this is on top of a couch.}
    \item \textit{A pillow on top of a white couch.}
    \item \textit{This is a couch and it is under a window.}
    \item \textit{This is a stool and it is on top of a rug.}
    \item \textit{This is a side table under a lamp.}
    \item \textit{This is a ceiling light next to the window.}
    \item \textit{This is an end table and it is below a lamp.}
    \item \textit{These are books and they are on the table.}
    \item \textit{This is a couch and it is in front of the wall.}
    \item \textit{White horizontal blinds in a well lit room.}
    \item \textit{This is a striped throw pillow on the loveseat.}
    \item \textit{The pillow is on top of the chair.}
    \item \textit{This is a window and it is next to a door.}
    \item \textit{This is a hurricane candle and it is on top of a cabinet.}
    \item \textit{This is a vase and it is on top of the table.}
    \item \textit{This is a vent in the ceiling.}
    \item \textit{This is a fish and it is on top of a cabinet.}
    \item \textit{This is a window behind a chair.}
    \item \textit{This is a trash can against a wall.}
    \item \textit{Two cream colored cushioned chairs with blue pillows adjacent to each other.}
\end{enumerate}

\noindent \textbf{Replica Dataset Affordance Queries for Office Scenes}:
\begin{enumerate}
    \item \textit{Something to watch the news on}
    \item \textit{Something to tell the time}
    \item \textit{Something comfortable to sit on}
    \item \textit{Something to dispose of wastepaper in}
    \item \textit{Something to add light into the room}
\end{enumerate}

\noindent \textbf{Replica Dataset Affordance Queries for Room Scenes}:
\begin{enumerate}
    \item \textit{Somewhere to store decorative cups}
    \item \textit{Something to add light into the room}
    \item \textit{Somewhere to set food for dinner}
    \item \textit{Something I can open with my keys}
    \item \textit{Somewhere to sit upright for a work call}
\end{enumerate}

\noindent \textbf{Replica Dataset Negation Queries for Office Scenes}:
\begin{enumerate}
    \item \textit{Something to sit on other than a chair}
    \item \textit{Something very heavy, unlike a clock}
    \item \textit{Something rigid, unlike a cushion}
    \item \textit{Something small, unlike a couch}
    \item \textit{Something light, unlike a table}
\end{enumerate}

\noindent \textbf{Replica Dataset Negation Queries for Room Scenes}:
\begin{enumerate}
    \item \textit{Something small, unlike a cabinet}
    \item \textit{Something light, unlike a table}
    \item \textit{Something soft, unlike a table}
    \item \textit{Something not transparent, unlike a window}
    \item \textit{Something rigid, unlike a rug}
\end{enumerate}

\noindent \textbf{REAL Lab Scan Descriptive queries}:

\begin{enumerate}
    \item \textit{A pair of red and white sneakers}
    \item \textit{A NASA t-shirt}
    \item \textit{A Rubik's cube}
    \item \textit{A basketball}
    \item \textit{A toy car}
    \item \textit{A backpack}
    \item \textit{An office chair}
    \item \textit{A pair of headphones}
    \item \textit{A yellow jacket}
    \item \textit{A laundry bag}
\end{enumerate}

\noindent \textbf{REAL Lab Affordance Queries:}
\begin{enumerate}
    \item \textit{Something to use to disassemble or take apart a laptop}
    \item \textit{Something to use for cooling a CPU}
    \item \textit{Something to use for carrying books day to day}
    \item \textit{Something to use for temporarily securing a broken zipper}
    \item \textit{Something to use to help a student understand how a computer works}
    \item \textit{An object that is used in a sport involving rims and nets}
    \item \textit{Something to keep myself from getting distracted by loud noises}
    \item \textit{Something to help explain math proofs to a student}
    \item \textit{Something I can use to protect myself from the harsh winter in Canada}
    \item \textit{Something fun to pass the time with}
\end{enumerate}

\noindent \textbf{REAL Lab Negation Queries:}
\begin{enumerate}
    \item \textit{A toy for someone who dislikes basketball}
    \item \textit{Shoes that you wouldn't wear to something formal}
    \item \textit{Something to protect me from the rain that's not an umbrella}
    \item \textit{Shoes that are not red and white}
    \item \textit{Something to make a cape with that's not green}
    \item \textit{Something to drink other than soda}
    \item \textit{Something to use for exercise other than weights}
    \item \textit{Something to wear unrelated to space or science}
    \item \textit{Something light to store belongings, not a backpack}
    \item \textit{Something to play with that's not a puzzle or colorful}
\end{enumerate}

\section{Navigation Experiments}
\label{app:sec:jackal}

For our navigation experiments with the Jackal robot. Our robot is equipped with a VLP-16 lidar and a foward-facing Realsense D435i camera. We begin by building a pointcloud of the \href{https://montrealrobotics.ca/}{REAL Lab} using the onboard VLP-16 and Open3d SLAM~\cite{jelavic2022open3d}. The initial Jackal pointcloud does not include task-relevant objects and is downprojected to a 2D costmap for navigation using the base Jackal ROS stack.

We then stage two separate scenes with different objects: one for object search and another for traversability estimation. In both cases, we map the scene with an Azure Kinect Camera and rely on~RTAB-Map~\cite{labbe2019rtab} to obtain camera poses and the scene point cloud. We proceed to build a \coolname~representation and register the scene point cloud with the initial Jackal map. For our navigation experiments, we only use the objects $\objset_T$.

For object search queries, we use the LLM Planner described in Section \ref{app:sec:llm-planner-details} as part of a simple state machine. The robot first attempts to go look at the 3D coordinates of the most relevant object identified in $\objset_T$ by the LLM Planner. We then pass the onboard camera image to LLaVA~\cite{llava} and ask if the target object is in view. If not, we remove the target object from the scene graph and ask the LLM Planner to provide a new likely location for the object in the scene with the following GPT-4 system prompt:

\begin{lstlisting}[language=gptprompt, caption=GPT system prompt for object localization., escapeinside={(*@}{@*)}]
(*@
\begin{minipage}{\linewidth}
The object described as `{description}' is not in the scene. Perhaps someone has moved it, or put it away. Let's try to find the object by visiting the likely places, storage or containers that are appropriate for the missing object (eg: a a cabinet for a wineglass, or closet for a broom). The new query is: find a likely container or storage space where someone typically would have moved the object described as `{description}'?
\end{minipage}
@*)
\end{lstlisting}

For traversability estimation, we task GPT to classify a given object as traversable or non-traversable based on its description and possible tags. The system prompt is:

\begin{lstlisting}[language=gptprompt, caption=GPT-4 system prompt for traversability estimation., escapeinside={(*@}{@*)}]
(*@
\begin{minipage}{\linewidth}
 You are a wheeled robot that can push a maximum of 5 pounds or 2.27 kg. Can you traverse through or push an object identified as `{description}' with possible tags `{possible\_tags}'? Specifically, is it possible for you to push the object out of its path without damaging yourself?
\end{minipage}
@*)
\end{lstlisting}

We then take the pointclouds of each non-traversable objects and downproject them in the Jackal costmap before launching the navigation episode. The goal is provided in this case as a specific pose in the room.

For all experiments in this section, we run a local instance of LLaVA offboard on a desktop when needed and otherwise use the GPT-4 API for LLM queries.

\section{Limitations}

As indicated in Sec.~\ref{sec:limitations}, there are a few failure modes of \coolname{} that remain to be addressed in subsequent work.
In particular, the LLaVA-7B~\cite{llava} model used for node captioning misclassifies a non-negligible number of small objects as \emph{toothbrushes} or \emph{pairs of scissors}.
We believe that using more performant vision-language models, including instruction-finetuned variants of LLaVA~\cite{llavarlhf} can alleviate this issue to a large extent.
This will, in turn, improve the node and edge precisions of 3D scene graphs beyond what we report in Table~\ref{tab:sg-accuracy}.

In this work, we do not explicitly focus on improving LLM-based planning over 3D scene graphs. We refer the interested reader to concurrent work, SayPlan~\cite{rana2023sayplan}, for insights into how one might leverage the hierarchy inherent in 3D scene graphs, for efficient planning.

\bibliographystyle{IEEEtran}
\bibliography{IEEEtranBST/IEEEabrv, root}

\begin{thebibliography}{10}
\providecommand{\url}[1]{#1}
\csname url@samestyle\endcsname
\providecommand{\newblock}{\relax}
\providecommand{\bibinfo}[2]{#2}
\providecommand{\BIBentrySTDinterwordspacing}{\spaceskip=0pt\relax}
\providecommand{\BIBentryALTinterwordstretchfactor}{4}
\providecommand{\BIBentryALTinterwordspacing}{\spaceskip=\fontdimen2\font plus
\BIBentryALTinterwordstretchfactor\fontdimen3\font minus \fontdimen4\font\relax}
\providecommand{\BIBforeignlanguage}[2]{{%
\expandafter\ifx\csname l@#1\endcsname\relax
\typeout{** WARNING: IEEEtran.bst: No hyphenation pattern has been}%
\typeout{** loaded for the language `#1'. Using the pattern for}%
\typeout{** the default language instead.}%
\else
\language=\csname l@#1\endcsname
\fi
#2}}
\providecommand{\BIBdecl}{\relax}
\BIBdecl

\bibitem{newcombe2011kinectfusion}
R.~A. Newcombe, S.~Izadi, O.~Hilliges, D.~Molyneaux, D.~Kim, A.~J. Davison, P.~Kohi, J.~Shotton, S.~Hodges, and A.~Fitzgibbon, ``Kinectfusion: Real-time dense surface mapping and tracking,'' in \emph{IEEE international symposium on mixed and augmented reality}.\hskip 1em plus 0.5em minus 0.4em\relax IEEE, 2011, pp. 127--136.

\bibitem{whelan2015elasticfusion}
T.~Whelan, S.~Leutenegger, R.~Salas-Moreno, B.~Glocker, and A.~Davison, ``Elasticfusion: Dense slam without a pose graph,'' in \emph{Robotics Science and Systems}, 2015.

\bibitem{wen2023bundlesdf}
B.~Wen, J.~Tremblay, V.~Blukis, S.~Tyree, T.~M{\"u}ller, A.~Evans, D.~Fox, J.~Kautz, and S.~Birchfield, ``{BundleSDF}: Neural 6-dof tracking and 3d reconstruction of unknown objects,'' in \emph{Proceedings of the IEEE/CVF Conference on Computer Vision and Pattern Recognition}, 2023, pp. 606--617.

\bibitem{sucar2021imap}
E.~Sucar, S.~Liu, J.~Ortiz, and A.~J. Davison, ``{iMAP}: Implicit mapping and positioning in real-time,'' in \emph{Proceedings of the IEEE/CVF International Conference on Computer Vision}, 2021, pp. 6229--6238.

\bibitem{zhu2022niceslam}
Z.~Zhu, S.~Peng, V.~Larsson, W.~Xu, H.~Bao, Z.~Cui, M.~R. Oswald, and M.~Pollefeys, ``Nice-slam: Neural implicit scalable encoding for slam,'' in \emph{Proceedings of the IEEE/CVF Conference on Computer Vision and Pattern Recognition}, 2022, pp. 12\,786--12\,796.

\bibitem{schonberger2016sfm}
J.~L. Schonberger and J.-M. Frahm, ``Structure-from-motion revisited,'' in \emph{Proceedings of the IEEE conference on computer vision and pattern recognition}, 2016, pp. 4104--4113.

\bibitem{schonberger2016pixelwise}
J.~L. Sch{\"o}nberger, E.~Zheng, J.-M. Frahm, and M.~Pollefeys, ``Pixelwise view selection for unstructured multi-view stereo,'' in \emph{European Conference on Computer Vision}.\hskip 1em plus 0.5em minus 0.4em\relax Springer, 2016, pp. 501--518.

\bibitem{mccormac2017semanticfusion}
J.~McCormac, A.~Handa, A.~Davison, and S.~Leutenegger, ``Semanticfusion: Dense 3d semantic mapping with convolutional neural networks,'' in \emph{IEEE International Conference on Robotics and automation (ICRA)}.\hskip 1em plus 0.5em minus 0.4em\relax IEEE, 2017, pp. 4628--4635.

\bibitem{runz2018maskfusion}
M.~Runz, M.~Buffier, and L.~Agapito, ``Maskfusion: Real-time recognition, tracking and reconstruction of multiple moving objects,'' in \emph{IEEE International Symposium on Mixed and Augmented Reality (ISMAR)}.\hskip 1em plus 0.5em minus 0.4em\relax IEEE, 2018, pp. 10--20.

\bibitem{mccormac2018fusion++}
J.~McCormac, R.~Clark, M.~Bloesch, A.~Davison, and S.~Leutenegger, ``Fusion++: Volumetric object-level slam,'' in \emph{international conference on 3D vision (3DV)}.\hskip 1em plus 0.5em minus 0.4em\relax IEEE, 2018, pp. 32--41.

\bibitem{narita2019panopticfusion}
G.~Narita, T.~Seno, T.~Ishikawa, and Y.~Kaji, ``Panopticfusion: Online volumetric semantic mapping at the level of stuff and things,'' in \emph{2019 IEEE/RSJ International Conference on Intelligent Robots and Systems (IROS)}.\hskip 1em plus 0.5em minus 0.4em\relax IEEE, 2019, pp. 4205--4212.

\bibitem{Qian2022pocd}
J.~Qian, V.~Chatrath, J.~Yang, J.~Servos, A.~P. Schoellig, and S.~L. Waslander, ``{POCD:} probabilistic object-level change detection and volumetric mapping in semi-static scenes,'' in \emph{Robotics Science and Systems}, K.~Hauser, D.~A. Shell, and S.~Huang, Eds., 2022.

\bibitem{qian2023povslam}
J.~Qian, V.~Chatrath, J.~Servos, A.~Mavrinac, W.~Burgard, S.~L. Waslander, and A.~P. Schoellig, ``{POV-SLAM:} probabilistic object-aware variational {SLAM} in semi-static environments,'' in \emph{Robotics Science and Systems}, K.~E. Bekris, K.~Hauser, S.~L. Herbert, and J.~Yu, Eds., 2023.

\bibitem{li2021odam}
K.~Li, D.~DeTone, Y.~F.~S. Chen, M.~Vo, I.~Reid, H.~Rezatofighi, C.~Sweeney, J.~Straub, and R.~Newcombe, ``{ODAM}: Object detection, association, and mapping using posed rgb video,'' in \emph{Proceedings of International Conference on Computer Vision}, 2021.

\bibitem{zins2022oaslam}
M.~Zins, G.~Simon, and M.-O. Berger, ``{OA-SLAM}: Leveraging objects for camera relocalization in visual slam,'' in \emph{IEEE International Symposium on Mixed and Augmented Reality (ISMAR)}.\hskip 1em plus 0.5em minus 0.4em\relax IEEE, 2022, pp. 720--728.

\bibitem{liu20233dovs}
K.~Liu, F.~Zhan, J.~Zhang, M.~Xu, Y.~Yu, A.~E. Saddik, C.~Theobalt, E.~Xing, and S.~Lu, ``3d open-vocabulary segmentation with foundation models,'' \emph{arXiv preprint arXiv:2305.14093}, 2023.

\bibitem{conceptfusion}
K.~M. Jatavallabhula, A.~Kuwajerwala, Q.~Gu, M.~Omama, G.~Iyer, S.~Saryazdi, T.~Chen, A.~Maalouf, S.~Li, N.~V. Keetha, A.~Tewari, J.~B. Tenenbaum, C.~M. de~Melo, K.~M. Krishna, L.~Paull, F.~Shkurti, and A.~Torralba, ``{ConceptFusion}: Open-set multimodal 3d mapping,'' in \emph{Robotics: Science and Systems}, 2023.

\bibitem{peng2023openscene}
S.~Peng, K.~Genova, C.~Jiang, A.~Tagliasacchi, M.~Pollefeys, T.~Funkhouser \emph{et~al.}, ``Openscene: 3d scene understanding with open vocabularies,'' in \emph{Proceedings of the IEEE/CVF Conference on Computer Vision and Pattern Recognition}, 2023, pp. 815--824.

\bibitem{ding2023lowis3d}
R.~Ding, J.~Yang, C.~Xue, W.~Zhang, S.~Bai, and X.~Qi, ``Lowis3d: Language-driven open-world instance-level 3d scene understanding,'' \emph{arXiv preprint arXiv:2308.00353}, 2023.

\bibitem{ding2023pla}
R.~Ding, J.~Yang, C.~Xue, W.~Zhang, S.~Bai, and X.~{Qi}, ``{PLA}: Language-driven open-vocabulary 3d scene understanding,'' in \emph{Proceedings of Computer Vision and Pattern Recognition}, 2023.

\bibitem{zhang2023clipfo3d}
J.~Zhang, R.~Dong, and K.~Ma, ``{CLIP-FO3D}: Learning free open-world 3d scene representations from 2d dense clip,'' \emph{arXiv preprint arXiv:2303.04748}, 2023.

\bibitem{tschernezki2022n3f}
V.~Tschernezki, I.~Laina, D.~Larlus, and A.~Vedaldi, ``Neural feature fusion fields: 3d distillation of self-supervised 2d image representations,'' in \emph{International Conference on 3D Vision (3DV)}.\hskip 1em plus 0.5em minus 0.4em\relax IEEE, 2022.

\bibitem{kobayashi2022ffd}
S.~Kobayashi, E.~Matsumoto, and V.~Sitzmann, ``Decomposing nerf for editing via feature field distillation,'' \emph{Neural Information Processing Systems}, vol.~35, pp. 23\,311--23\,330, 2022.

\bibitem{clipfields}
N.~M.~M. Shafiullah, C.~Paxton, L.~Pinto, S.~Chintala, and A.~Szlam, ``Clip-fields: Weakly supervised semantic fields for robotic memory,'' in \emph{Robotics: Science and Systems}, K.~E. Bekris, K.~Hauser, S.~L. Herbert, and J.~Yu, Eds., 2023.

\bibitem{tsagkas2023vlfields}
N.~Tsagkas, O.~Mac~Aodha, and C.~X. Lu, ``Vl-fields: Towards language-grounded neural implicit spatial representations,'' \emph{arXiv preprint arXiv:2305.12427}, 2023.

\bibitem{lerf2023}
J.~Kerr, C.~M. Kim, K.~Goldberg, A.~Kanazawa, and M.~Tancik, ``{LERF}: Language embedded radiance fields,'' in \emph{International Conference on Computer Vision (ICCV)}, 2023.

\bibitem{huang23avlmaps}
C.~Huang, O.~Mees, A.~Zeng, and W.~Burgard, ``Audio visual language maps for robot navigation,'' \emph{arXiv preprint arXiv:2303.07522}, 2023.

\bibitem{shen2023distilled}
\BIBentryALTinterwordspacing
W.~Shen, G.~Yang, A.~Yu, J.~Wong, L.~P. Kaelbling, and P.~Isola, ``Distilled feature fields enable few-shot manipulation,'' in \emph{International Conference on Robot Learning}, 2023. [Online]. Available: \url{https://openreview.net/forum?id=Rb0nGIt_kh5}
\BIBentrySTDinterwordspacing

\bibitem{Engelmann2023openreno}
F.~Engelmann, F.~Manhardt, M.~Niemeyer, K.~Tateno, M.~Pollefeys, and F.~Tombari, ``Open-set 3d scene segmentation with rendered novel views,'' 2023.

\bibitem{mazur2023feature}
K.~Mazur, E.~Sucar, and A.~J. Davison, ``Feature-realistic neural fusion for real-time, open set scene understanding,'' in \emph{IEEE International Conference on Robotics and Automation (ICRA)}.\hskip 1em plus 0.5em minus 0.4em\relax IEEE, 2023.

\bibitem{clip}
A.~Radford, J.~W. Kim, C.~Hallacy, A.~Ramesh, G.~Goh, S.~Agarwal, G.~Sastry, A.~Askell, P.~Mishkin, J.~Clark \emph{et~al.}, ``Learning transferable visual models from natural language supervision,'' in \emph{International Conference on Machine Learning}.\hskip 1em plus 0.5em minus 0.4em\relax PMLR, 2021.

\bibitem{gpt4}
OpenAI, ``Gpt-4 technical report,'' \emph{arXiv preprint arXiv:2303.08774}, 2023.

\bibitem{kirillov2023sam}
A.~Kirillov, E.~Mintun, N.~Ravi, H.~Mao, C.~Rolland, L.~Gustafson, T.~Xiao, S.~Whitehead, A.~C. Berg, W.-Y. Lo \emph{et~al.}, ``Segment anything,'' \emph{Proceedings of International Conference on Computer Vision}, 2023.

\bibitem{liu2023grounding}
S.~Liu, Z.~Zeng, T.~Ren, F.~Li, H.~Zhang, J.~Yang, C.~Li, J.~Yang, H.~Su, J.~Zhu \emph{et~al.}, ``Grounding dino: Marrying dino with grounded pre-training for open-set object detection,'' \emph{arXiv preprint arXiv:2303.05499}, 2023.

\bibitem{stablediffusion}
R.~Rombach, A.~Blattmann, D.~Lorenz, P.~Esser, and B.~Ommer, ``High-resolution image synthesis with latent diffusion models,'' in \emph{Proceedings of the IEEE/CVF conference on computer vision and pattern recognition}, 2022, pp. 10\,684--10\,695.

\bibitem{hong20223d}
Y.~Hong, Y.~Du, C.~Lin, J.~Tenenbaum, and C.~Gan, ``3d concept grounding on neural fields,'' \emph{Neural Information Processing Systems}, 2022.

\bibitem{hong20233d}
Y.~Hong, C.~Lin, Y.~Du, Z.~Chen, J.~B. Tenenbaum, and C.~Gan, ``3d concept learning and reasoning from multi-view images,'' in \emph{Proceedings of Computer Vision and Pattern Recognition}, 2023.

\bibitem{hong20233d-llm}
Y.~Hong, H.~Zhen, P.~Chen, S.~Zheng, Y.~Du, Z.~Chen, and C.~Gan, ``3d-llm: Injecting the 3d world into large language models,'' \emph{Neural Information Processing Systems}, 2023.

\bibitem{shridar2021cliport}
M.~Shridhar, L.~Manuelli, and D.~Fox, ``{CLIPort}: What and where pathways for robotic manipulation,'' in \emph{Conference on Robot Learning}, vol. 164.\hskip 1em plus 0.5em minus 0.4em\relax {PMLR}, 2021, pp. 894--906.

\bibitem{sharma2023llerftogo}
\BIBentryALTinterwordspacing
S.~Sharma, A.~Rashid, C.~M. Kim, J.~Kerr, L.~Y. Chen, A.~Kanazawa, and K.~Goldberg, ``Language embedded radiance fields for zero-shot task-oriented grasping,'' in \emph{International Conference on Robot Learning}, 2023. [Online]. Available: \url{https://openreview.net/forum?id=k-Fg8JDQmc}
\BIBentrySTDinterwordspacing

\bibitem{gadre2022cow}
S.~Y. Gadre, M.~Wortsman, G.~Ilharco, L.~Schmidt, and S.~Song, ``Clip on wheels: Zero-shot object navigation as object localization and exploration,'' \emph{arXiv preprint arXiv:2203.10421}, 2022.

\bibitem{shah2022lmnav}
D.~Shah, B.~Osi{\'n}ski, S.~Levine \emph{et~al.}, ``Lm-nav: Robotic navigation with large pre-trained models of language, vision, and action,'' in \emph{International Conference on Robot Learning}.\hskip 1em plus 0.5em minus 0.4em\relax PMLR, 2023.

\bibitem{Fisher2011scenegraph}
M.~Fisher, M.~Savva, and P.~Hanrahan, ``Characterizing structural relationships in scenes using graph kernels,'' \emph{{ACM} Trans. Graph.}, vol.~30, no.~4, p.~34, 2011.

\bibitem{gay2019visual}
P.~Gay, J.~Stuart, and A.~Del~Bue, ``Visual graphs from motion (vgfm): Scene understanding with object geometry reasoning,'' in \emph{Asian Conference on Computer Vision}.\hskip 1em plus 0.5em minus 0.4em\relax Springer, 2019.

\bibitem{Armeni20193DSG}
I.~Armeni, Z.-Y. He, J.~Gwak, A.~R. Zamir, M.~Fischer, J.~Malik, and S.~Savarese, ``3d scene graph: A structure for unified semantics, 3d space, and camera,'' in \emph{Proceedings of International Conference on Computer Vision}, October 2019.

\bibitem{kim20193dsg}
U.-H. Kim, J.-M. Park, T.-J. Song, and J.-H. Kim, ``3-d scene graph: A sparse and semantic representation of physical environments for intelligent agents,'' \emph{IEEE transactions on cybernetics}, vol.~50, no.~12, pp. 4921--4933, 2019.

\bibitem{wald2020learning}
J.~Wald, H.~Dhamo, N.~Navab, and F.~Tombari, ``Learning 3d semantic scene graphs from 3d indoor reconstructions,'' in \emph{Proceedings of Computer Vision and Pattern Recognition}, 2020.

\bibitem{rosinol2021kimera}
A.~Rosinol, A.~Violette, M.~Abate, N.~Hughes, Y.~Chang, J.~Shi, A.~Gupta, and L.~Carlone, ``Kimera: From slam to spatial perception with 3d dynamic scene graphs,'' \emph{The International Journal of Robotics Research}, vol.~40, no. 12-14, pp. 1510--1546, 2021.

\bibitem{hughes2022hydra}
N.~Hughes, Y.~Chang, and L.~Carlone, ``Hydra: A real-time spatial perception system for 3d scene graph construction and optimization,'' \emph{arXiv preprint arXiv:2201.13360}, 2022.

\bibitem{wu2021scenegraphfusion}
S.-C. Wu, J.~Wald, K.~Tateno, N.~Navab, and F.~Tombari, ``Scenegraphfusion: Incremental 3d scene graph prediction from rgb-d sequences,'' in \emph{Proceedings of Computer Vision and Pattern Recognition}, 2021.

\bibitem{agia2022taskography}
C.~Agia, K.~M. Jatavallabhula, M.~Khodeir, O.~Miksik, V.~Vineet, M.~Mukadam, L.~Paull, and F.~Shkurti, ``Taskography: Evaluating robot task planning over large 3d scene graphs,'' in \emph{International Conference on Robot Learning}.\hskip 1em plus 0.5em minus 0.4em\relax PMLR, 2022.

\bibitem{rana2023sayplan}
\BIBentryALTinterwordspacing
K.~Rana, J.~Abou-Chakra, S.~Garg, J.~Haviland, I.~Reid, and N.~Suenderhauf, ``Sayplan: Grounding large language models using 3d scene graphs for scalable task planning,'' in \emph{International Conference on Robot Learning}, 2023. [Online]. Available: \url{https://openreview.net/forum?id=wMpOMO0Ss7a}
\BIBentrySTDinterwordspacing

\bibitem{oquab2023dinov2}
M.~Oquab, T.~Darcet, T.~Moutakanni, H.~Vo, M.~Szafraniec, V.~Khalidov, P.~Fernandez, D.~Haziza, F.~Massa, A.~El-Nouby \emph{et~al.}, ``Dinov2: Learning robust visual features without supervision,'' \emph{arXiv preprint arXiv:2304.07193}, 2023.

\bibitem{zhang2023ram}
Y.~Zhang, X.~Huang, J.~Ma, Z.~Li, Z.~Luo, Y.~Xie, Y.~Qin, T.~Luo, Y.~Li, S.~Liu \emph{et~al.}, ``Recognize anything: A strong image tagging model,'' \emph{arXiv preprint arXiv:2306.03514}, 2023.

\bibitem{llava}
H.~Liu, C.~Li, Q.~Wu, and Y.~J. Lee, ``Visual instruction tuning,'' \emph{arXiv preprint arXiv:2304.08485}, 2023.

\bibitem{replica}
J.~Straub, T.~Whelan, L.~Ma, Y.~Chen, E.~Wijmans, S.~Green, J.~J. Engel, R.~Mur-Artal, C.~Ren, S.~Verma, A.~Clarkson, M.~Yan, B.~Budge, Y.~Yan, X.~Pan, J.~Yon, Y.~Zou, K.~Leon, N.~Carter, J.~Briales, T.~Gillingham, E.~Mueggler, L.~Pesqueira, M.~Savva, D.~Batra, H.~M. Strasdat, R.~D. Nardi, M.~Goesele, S.~Lovegrove, and R.~Newcombe, ``The {R}eplica dataset: A digital replica of indoor spaces,'' \emph{arXiv preprint arXiv:1906.05797}, 2019.

\bibitem{clipseg}
T.~L{\"u}ddecke and A.~Ecker, ``Image segmentation using text and image prompts,'' in \emph{Proceedings of the IEEE/CVF Conference on Computer Vision and Pattern Recognition}, 2022, pp. 7086--7096.

\bibitem{lseg}
B.~Li, K.~Q. Weinberger, S.~Belongie, V.~Koltun, and R.~Ranftl, ``Language-driven semantic segmentation,'' in \emph{International Conference on Learning Representations}, 2022.

\bibitem{openseg}
G.~Ghiasi, X.~Gu, Y.~Cui, and T.-Y. Lin, ``Scaling open-vocabulary image segmentation with image-level labels,'' in \emph{European Conference on Computer Vision}.\hskip 1em plus 0.5em minus 0.4em\relax Springer, 2022, pp. 540--557.

\bibitem{maskclip}
C.~Zhou, C.~C. Loy, and B.~Dai, ``Extract free dense labels from clip,'' in \emph{European Conference on Computer Vision (ECCV)}, 2022.

\bibitem{du2020compositional}
Y.~Du, S.~Li, and I.~Mordatch, ``Compositional visual generation with energy based models,'' in \emph{Neural Information Processing Systems}, 2020.

\bibitem{levine2023learning}
S.~Levine and D.~Shah, ``Learning robotic navigation from experience: principles, methods and recent results,'' \emph{Philosophical Transactions of the Royal Society B}, vol. 378, no. 1869, p. 20210447, 2023.

\bibitem{ai2thor}
E.~Kolve, R.~Mottaghi, W.~Han, E.~VanderBilt, L.~Weihs, A.~Herrasti, M.~Deitke, K.~Ehsani, D.~Gordon, Y.~Zhu \emph{et~al.}, ``Ai2-thor: An interactive 3d environment for visual ai,'' \emph{arXiv preprint arXiv:1712.05474}, 2017.

\bibitem{procthor}
M.~Deitke, E.~VanderBilt, A.~Herrasti, L.~Weihs, K.~Ehsani, J.~Salvador, W.~Han, E.~Kolve, A.~Kembhavi, and R.~Mottaghi, ``Procthor: Large-scale embodied ai using procedural generation,'' \emph{Advances in Neural Information Processing Systems}, vol.~35, pp. 5982--5994, 2022.

\bibitem{takmaz2023openmask3d}
A.~Takmaz, E.~Fedele, R.~W. Sumner, M.~Pollefeys, F.~Tombari, and F.~Engelmann, ``Openmask3d: Open-vocabulary 3d instance segmentation,'' \emph{arXiv preprint arXiv:2306.13631}, 2023.

\bibitem{lu2023ovir3d}
\BIBentryALTinterwordspacing
S.~Lu, H.~Chang, E.~P. Jing, A.~Boularias, and K.~Bekris, ``{OVIR}-3d: Open-vocabulary 3d instance retrieval without training on 3d data,'' in \emph{International Conference on Robot Learning}, 2023. [Online]. Available: \url{https://openreview.net/forum?id=gVBvtRqU1_}
\BIBentrySTDinterwordspacing

\bibitem{chang2023ogsv}
\BIBentryALTinterwordspacing
H.~Chang, K.~Boyalakuntla, S.~Lu, S.~Cai, E.~P. Jing, S.~Keskar, S.~Geng, A.~Abbas, L.~Zhou, K.~Bekris, and A.~Boularious, ``Context-aware entity grounding with open-vocabulary 3d scene graphs,'' in \emph{International Conference on Robot Learning}, 2023. [Online]. Available: \url{https://openreview.net/forum?id=cjEI5qXoT0}
\BIBentrySTDinterwordspacing

\bibitem{jelavic2022open3d}
E.~Jelavic, J.~Nubert, and M.~Hutter, ``Open3d slam: Point cloud based mapping and localization for education,'' in \emph{Robotic Perception and Mapping: Emerging Techniques, ICRA 2022 Workshop}.\hskip 1em plus 0.5em minus 0.4em\relax ETH Zurich, Robotic Systems Lab, 2022, p.~24.

\bibitem{labbe2019rtab}
M.~Labb{\'e} and F.~Michaud, ``Rtab-map as an open-source lidar and visual simultaneous localization and mapping library for large-scale and long-term online operation,'' \emph{Journal of Field Robotics}, vol.~36, no.~2, pp. 416--446, 2019.

\bibitem{llavarlhf}
Z.~Sun, S.~Shen, S.~Cao, H.~Liu, C.~Li, Y.~Shen, C.~Gan, L.-Y. Gui, Y.-X. Wang, Y.~Yang, K.~Keutzer, and T.~Darrell, ``Aligning large multimodal models with factually augmented rlhf,'' 2023.

\end{thebibliography}

\end{document}